\pdfoutput=1

\documentclass[11pt]{article}

\usepackage{acl}

\usepackage{times}
\usepackage{latexsym}

\usepackage[T1]{fontenc}

\usepackage[utf8]{inputenc}

\usepackage{microtype}
\usepackage{amsmath,bm}
\usepackage{amssymb}
\usepackage{bbm}
\usepackage{graphicx}
\usepackage{algorithm}
\usepackage{algpseudocode}
\usepackage{hyperref}
\usepackage{multirow}
\usepackage{booktabs}
\usepackage{subfigure}
\usepackage{float}
\usepackage{stfloats}

\usepackage{xcolor}

\usepackage[inline]{enumitem}

\title{TART: Improved Few-shot Text Classification Using \\Task-Adaptive Reference Transformation}

\author{
Shuo Lei{$^\dag$}, Xuchao Zhang{$^\ddag$}, Jianfeng He{$^\dag$},
Fanglan Chen{$^\dag$}, Chang-Tien Lu{$^\dag$}
\\ 
{$^\dag$}Department of Computer Science, Virginia Tech, Falls Church, VA, USA
\\{$^\ddag$}Microsoft, Redmond, WA, USA \\
\{\tt{slei,jianfenghe,fanglanc,ctlu}\}@vt.edu \\
\tt{xuchaozhang@microsoft.com}
 }

\begin{document}
\maketitle
\begin{abstract}
Meta-learning has emerged as a trending technique to tackle few-shot text classiﬁcation and achieve state-of-the-art performance. 
However, the performance of existing approaches heavily depends on the inter-class variance of the support set. As a result, it can perform well on tasks when the semantics of sampled classes are distinct while failing to differentiate classes with similar semantics.
In this paper, we propose a novel \textbf{T}ask-\textbf{A}daptive \textbf{R}eference \textbf{T}ransformation (TART) network, aiming to enhance the generalization by transforming the class prototypes to per-class fixed reference points in task-adaptive metric spaces.
To further maximize divergence between transformed prototypes in task-adaptive metric spaces, TART introduces a discriminative reference regularization among transformed prototypes.
Extensive experiments are conducted on four benchmark datasets and our method demonstrates clear superiority over the state-of-the-art models in all the datasets. In particular, our model surpasses the state-of-the-art method by 7.4\% and 5.4\% in 1-shot and 5-shot classification on the 20 Newsgroups dataset, respectively. Our code is available at \url{https://github.com/slei109/TART}
\end{abstract}

\section{Introduction}
Deep learning has achieved great success in many fields but a deficiency of supervised data is often experienced in real-world NLP applications. 
Few-shot text classification aims to perform classification with a limited number of training instances, which is crucial for many applications but remains to be a challenging task.

\begin{figure}[tb]
    \centering
    \includegraphics[trim=0cm 6.2cm 0cm 0cm, width=\linewidth]{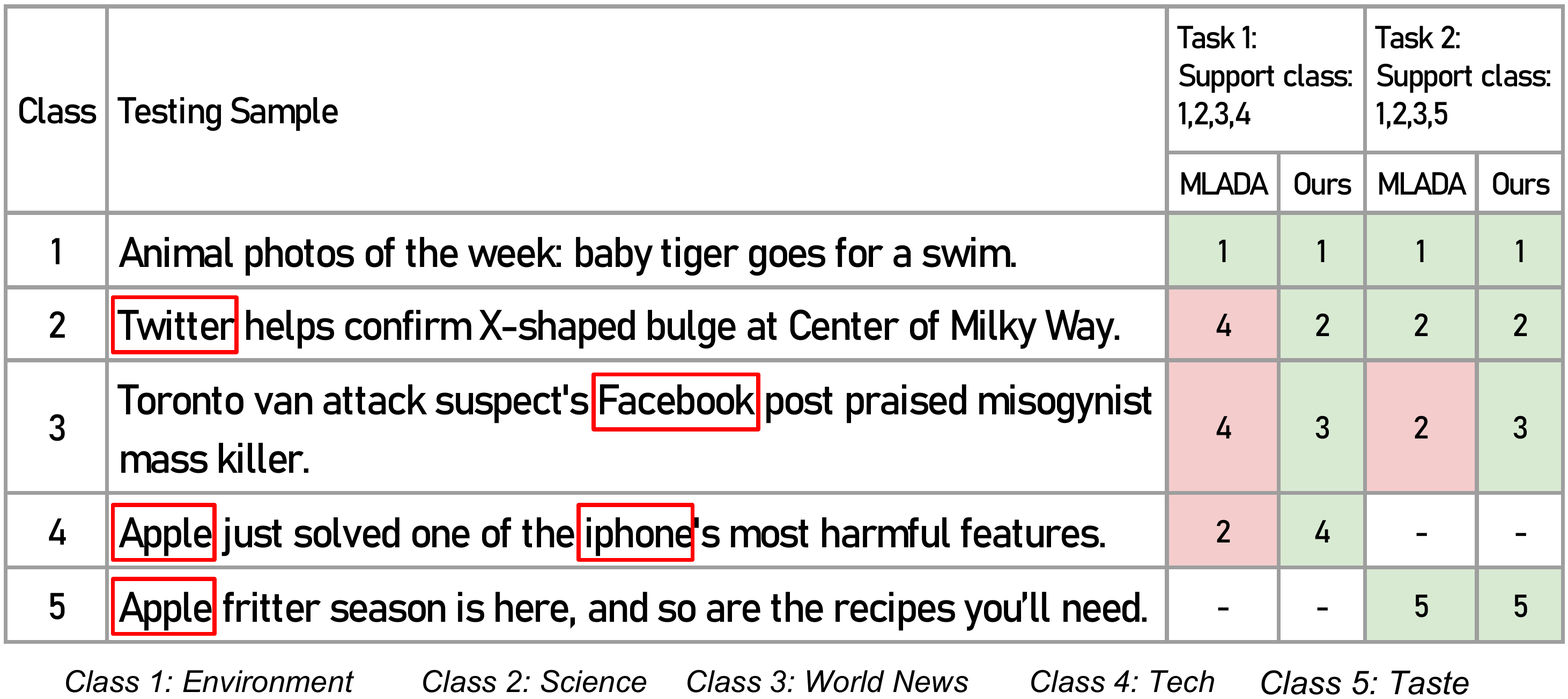}
    \caption{Prediction results of example tasks with different inter-class variance on the Huffpost dataset.
    MLADA~\cite{han2021meta} performs well on the task with high inter-class variance (e.g., Task 2: \textit{Environment}, \textit{Science}, \textit{World News}, \textit{Taste}), while it fails to distinguish the samples from a task with low inter-class variance
    (e.g., Task 1: \textit{Environment}, \textit{Science}, \textit{World News}, \textit{Tech}).}
    \label{fig:intro}
    \vspace{-0.3cm}
\end{figure}

Existing approaches for few-shot text classification mainly fall into two categories: i) prompt-based learning~\cite{brown2020language,gao-etal-2021-making,wang2021transprompt}, which utilizes Pre-trained Language Models (PLMs) to generate a textual answer in response to a given prompt.
Although producing promising results, these methods suffer from (1) requiring a large PLM to function properly; and (2) favoring certain issues which can be naturally posed as a “ﬁll-in-the-blank” problem and do not contain many output classes, rendering them inapplicable in many real-world scenarios. For instance, it is hard to run the large-scale model on devices with limited computing resources, like mobile devices. 
ii) meta-learning~\cite{finn2017model,snell2017prototypical}, also known as “\textit{learning to learn}”:
it improves the model's capacity to learn over multiple training tasks, allowing it to quickly adapt to new tasks with only a few training instances.
Since meta-learning-based methods ~\cite{gao2019hybrid,bao2019few,han2021meta} rely on learned cross-task transferable knowledge rather than recalling pre-trained knowledge gained through PLMs, these methods have no constraints on the target problem and are broadly studied on the small-scale model, making them more applicable to real-world applications. 

Despite the extraordinary effectiveness, we notice that current meta-learning-based approaches may have several limitations. For those methods that learn to represent each class independently in one feature space~\cite{snell2017prototypical,gao2019hybrid,han2021meta}, their performance is heavily dependent on the inter-class variance of the support set. Specifically, they address the overfitting issue in few-shot learning by directly adopting the hidden features of support samples as a classifier. Thus, they can perform well on tasks when the sampled classes are distinct while failing to differentiate classes with similar semantics.
As illustrated in Figure~\ref{fig:intro},
MLADA~\cite{han2021meta}, which leads to state-of-the-art performance, misclassifies the testing samples of \textit{Science}, \textit{World News} and \textit{Tech} during the testing stage, mostly because \textit{Science} and \textit{Tech} are similar and all three samples contain technology companies, which are difficult to distinguish. 
If we substitute the support class \textit{Science} with \textit{Taste}, which has clearly different semantic from \textit{Tech}, it can recognize all testing samples except the third one.
This example indicates that ignoring task-specific features and treating all tasks identically is inadequate. 
It is essential to consider the inter-class variance of support sets, particularly when annotated data is scarce.
Recently, ~\citet{bao2019few} leveraged distributional signatures to estimate class-specific word importance.
However, it requires extracting relevant statistics of each word from the source pool and the support set for each task, which is time-consuming. 
A natural question arises: \textit{how can we design an efficient method capable of capturing both cross-task transferable knowledge and task-specific features to enhance the model's generalization ability?}

\begin{figure}[tb]
    \centering
    \includegraphics[width=\linewidth]{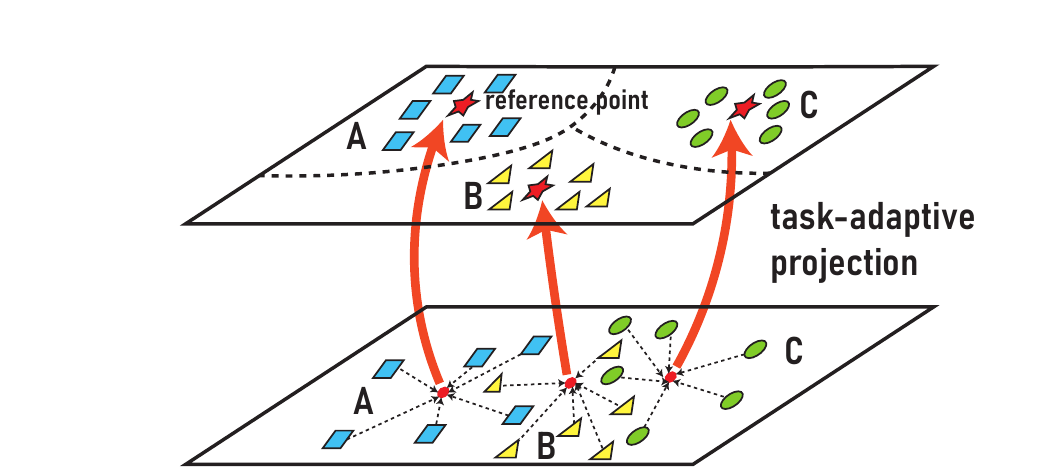}
    \caption{Feature representations in original metric space (bottom) and task-adaptive metric space (top).}
    \label{fig:metric}
    \vspace{-0.5cm}
\end{figure}

To tackle these issues, we resort to constructing a task-adaptive metric space via the meta-learning framework.
Figure~\ref{fig:metric} presents the key idea of the proposed method.
Intuitively, for comparable classes that cannot be distinguished in the original feature space, if we can project their class prototypes to per-class fixed points, referred to as reference points, in another small space, it is helpful to enhance the divergence between class prototypes in the transformed space.
Consequently, we propose a novel Task-Adaptive Reference Transfer Module that uses a linear transformation matrix to project embedding features into a task-specific metric space.
In addition, we design a discriminative reference regularization to maximize the distance between transformed prototypes in the task-adaptive metric space for each task.
We find the proposed method promotes the model to learn discriminative reference vectors and construct a stable metric space.

Our key contributions can be summarized as follows. 
1) We propose a \textbf{T}ask-\textbf{A}daptive \textbf{R}eference \textbf{T}ransformation (TART) network for few-shot text classification. 
The model enhances the generalization by transforming the class prototypes to per-class fixed reference points in task-adaptive metric spaces.
2) We propose a novel discriminative reference regularization to maximize divergence between transformed prototypes in task-adaptive metric spaces to further improve the performance.
3) We evaluate the proposed model on four popular datasets for few-shot text classification. 
Comprehensive experiments demonstrate that our TART consistently outperforms all the baselines for both 1-shot and 5-shot classification tasks. 
For instance, our model outperforms MLADA~\cite{han2021meta} model by 7.4\% and 5.4\% in 1-shot and 5-shot classification on the 20 Newsgroups dataset, respectively.

\section{Related Work}\label{sec:related}

\textbf{Few-shot learning}
Few-shot learning aims to learn a new concept representation from only a few annotated examples. Most existing works can be categorized into three groups: (1) Gradient-based meta-learners, including MAML~\cite{finn2017model}, MAML++~\cite{antoniou2018train}, and MetaNets~\cite{munkhdalai2017meta}. The prominent idea is to learn a proper initialization of the neural network, one can expect the network to adapt to novel tasks via backpropagation from limited samples.
(2) Graph neural network~\cite{garcia2017few,liu2019learning} based methods, which cast few-shot learning as a supervised message passing task and utilize graph neural networks to train it end-to-end.
(3) Metric-based methods~\cite{vinyals2016matching,snell2017prototypical,sung2018learning}, which aim to optimize the transferable embedding using metric learning approaches. 
Specifically, Matching networks~\cite{vinyals2016matching} learns sample-wise metric, where distances to samples are used to determine the label of the query.
Prototypical Networks~\cite{snell2017prototypical} extends the idea from samples
to class-wise metric, where all the samples of a specific class are grouped and considered as class prototypes. Then the prototypes are subsequently used for inference.

\noindent\textbf{Transfer learning and Prompt learning for PLMs}
Few-shot text classification relates closely to transfer learning~\cite{zhuang2020comprehensive} that aims to leverage knowledge from source domains to target domains.
Fine-tuning Pre-trained Language Models (PLMs)~\cite{devlin2018bert, raffel2020exploring, brown2020language, lei-etal-2022-uncertainty} can also be viewed as a type of transfer learning.
Recently,~\citet{gao-etal-2021-making} proposed a prompt-based approach to fine-tune PLMs in a few-shot learning setting for \textit{similar} tasks, which adapts PLMs to producing specific tokens corresponding to each class, instead of learning the prediction head. 
Meta-learning deviates from these settings by learning to quickly adapt the model to \textit{different} tasks with little training data available~\cite{wang2021transprompt}, typically formulated as a $N$-way $K$-shot problem. 

\textbf{Few-shot text classification}
Few-shot text classification has gained increasing attention in recent years.
\citet{yu2018diverse} used an adaptive metric learning approach to select an optimal distance metric for different tasks.
Induction Network~\cite{geng2019induction} aims to learn an appropriate distance metric to compare validation points with training points and make predictions through matching training points.
DMIN~\cite{geng2020dynamic} utilizes dynamic routing to provide more flexibility to memory-based few-shot learning in order to adapt the support sets better. 
~\citet{bao2019few} leveraged distributional signatures (e.g. word frequency and information entropy) to train a model within a meta-learning framework.
Another group of methods is to improve performance with the help of additional knowledge, including pre-trained text paraphrasing model~\cite{dopierre2021protaugment,chen2022contrastnet} and class-label semantic information~\cite{luo2021don}.
Recently,~\citet{hong2022lea} constructed a meta-level attention aspects dictionary and determined the top-$k$ most relevant attention aspects to utilize pre-trained models in few-shot learning. MLADA~\cite{han2021meta} is an adversarial network, which improves the domain adaptation ability of meta-learning.
However, none of these methods consider task-specific features, which is a key factor for few-shot text classification.


\section{Model}
In this section, we initially discuss the problem setting of few-shot text classification. Then, the overview of the proposed TART is presented in Section \ref{sec:overview}. The technical details for the Task-Adaptive Reference Transfer Module and Discriminative Reference Regularization are described in Sections \ref{sec:adapt} and \ref{sec:disc}, respectively.

\subsection{Problem Setting}\label{sec:prob}
In $N$-way $K$-shot text classification, the objective is to train a model $f_{\theta}(\cdot)$ that can classify a given query example using the support set $\mathcal{S}$, which comprises $K$ examples for each of the $N$ different classes considered.
Note that $f_{\theta}(\cdot)$ has \textit{not} been pre-trained on any large datasets in advance.
In accordance with prior works~\cite{bao2019few,han2021meta}, we use the episode training and testing protocols on account of their effectiveness. 

Consider that we are given texts from two non-overlapping sets of classes $\mathcal{C}_{train}$ and $\mathcal{C}_{test}$, i.e., $\mathcal{C}_{train}\cap \mathcal{C}_{test} = \emptyset$. 
The training set $\mathcal{D}_{train}$ is constructed from $\mathcal{C}_{train}$, whereas the test set $\mathcal{D}_{test}$ is derived from $\mathcal{C}_{test}$. The model $f_{\theta}(\cdot)$ is trained on $\mathcal{D}_{train}$ and evaluated on $\mathcal{D}_{test}$.
Both the training set $\mathcal{D}_{train}$ and testing set $\mathcal{D}_{test}$ are comprised of multiple episodes. 
Each episode consists of a support set $\mathcal{S} =\{(\bm{x}_i,y_{i})\}_{i=1}^{N\times K}$ and a query set $\mathcal{Q} = \{(\bm{x}_{j},y_{j})\}_{j=1}^{Q}$, where $\bm{x}$ represents a text, $y$ is a corresponding class label and $Q$ is the number of query samples.
Due to the fact that each episode comprises distinct classes, the model is trained to generalize effectively to few-shot scenarios.
After meta-training is completed, we evaluate the performance of its few-shot text classification on the test set $\mathcal{D}_{test}$ over all the episodes. For better understanding, we denote ``episode" as ``task" in the following context.

\begin{figure*}[tb]
    \centering
    \includegraphics[trim=0cm 5.7cm 1cm 5.5cm, width=0.97\linewidth]{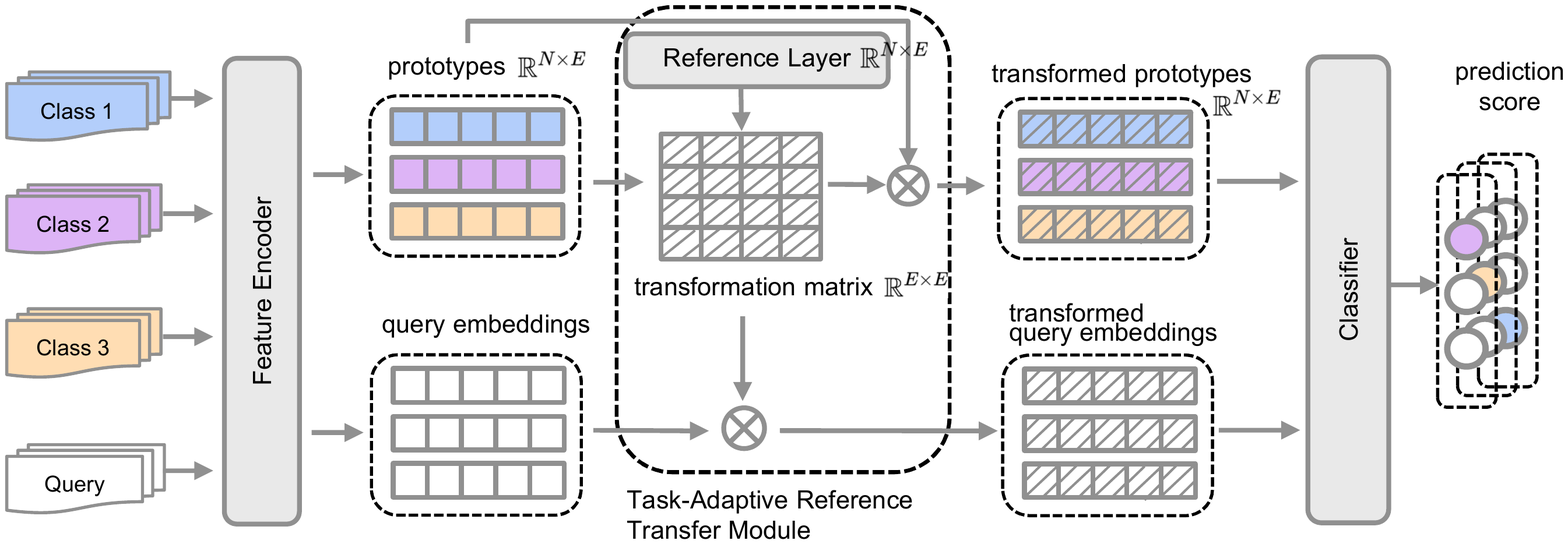}
    \vspace{-0.2cm}
    \caption{Illustration of the pipeline of TART for a 3-way 3-shot task with three query examples. After obtaining the embeddings of support and query inputs, the Task-Adaptive Reference Transfer Module is introduced to transform the embedding features into task-specific ones with a linear transformation matrix. Then, the query texts are classified by measuring the distance between each transformed prototype and transformed query embeddings in the task-adaptive metric space.}
    \label{fig:model}
\end{figure*}

\subsection{Overview}\label{sec:overview}
In this work, we resort to constructing a task-adaptive metric space to boost the performance of few-shot text classification. 
In contrast to previous approaches that construct the metric space using task-agnostic features, we propose to construct a task-adaptive metric space that enlarges the relative differences among sampled classes within a task.
Figure~\ref{fig:model} illustrates an overview of the proposed TART model.
Using a shared feature extractor, we encode contextual embeddings of the support and query texts for each episode.
Each class prototype is produced by averaging the support contextual embeddings.
Then, we offer a novel module, dubbed Task-Adaptive Reference Transfer Module, to construct a task-adaptive metric space and project contextual embeddings from the task-agnostic space to task-specific ones.
The classification of the query texts is accomplished by assigning each text the category of the closest prototype in the newly generated task-specific metric space. 
To learn discriminative reference vectors and construct a stable metric space,  we also propose Discriminative Reference Regularization (DRR), which measures the distance between transformed prototypes in the task-adaptive metric space. 



\subsection{Task-Adaptive Reference Transfer Module}\label{sec:adapt}
The key idea of the Task-Adaptive Reference Transfer Module is to acquire a feature transformer to construct a task-adaptive metric space.
Intuitively, for comparable classes that cannot be distinguished in the original feature space, if we can project their class prototypes to per-class fixed points, referred to as reference points, in another small space, it is helpful to enhance the divergence between class prototypes in the transformed space.
Below, we describe how to construct a task-adaptive metric space and make a classification based on it.


Different from learning a non-linear transformation matrix directly, our model adopts a linear transformation matrix calculated by using the reference layer and the prototypes of the support set.
This can effectively avoid overfitting since it introduces fewer learnable parameters.
First, we introduce a set of reference vectors $\{\bm{r}_1,\dots,\bm{r}_N\}$ as the fixed points for the transformed space, which are learned via a linear layer, dubbed reference layer.
We use the weight matrix of the reference layer and the prototype set of the support contextual embedding to compute the transformation matrix. Formally, let $R$ represent the weight matrix of the reference layer and $P$ denote the prototype matrix of the support texts. We construct the transformation matrix $W$ by ﬁnding a matrix such that $PW=R$.

Specifically, given a $N$-way $K$-shot episode (task), each class prototype is obtained by averaging the support contextual embeddings:
\begin{align}
\bm{p}_c=\frac{1}{|\mathcal{S}_c|}\sum_{(\bm{x}_i,y_i)\in \mathcal{S}_c}f_{\theta}(\bm{x}_i),
\end{align}
where $S_c$ denotes the support samples for the class $c$.
Accordingly, the reference weight matrix $R$ is defined as $[\frac{\bm{r}_1}{\parallel \bm{r}_1\parallel},\dots,\frac{\bm{r}_{N}}{\parallel \bm{r}_{N}\parallel}]$, where $R \in \mathbb{R}^{N \times E}$.
Note that each row in $R$ is the per-class reference vector and is learned during the training stage.
In general, $P$ is a non-square matrix and we can calculate its generalized inverse~\cite{ben2003generalized} with $P^{+} = \{P^{T}P\}^{-1}P^{T}$. Thus, the transformation matrix is computed as $W=P^{+}R$, where $W\in \mathbb{R}^{E\times E}$.

For each query input, we calculate the probability of belonging to class $c$ by applying a softmax function over the distance between the transformed query embeddings and each transformed prototype in the task-adaptive metric space. Concretely, given a distance function $d$, for each query input $\bm{x}_q$ and prototype set $\mathcal{P}=\{\bm{p}_1,\dots,\bm{p}_{N}\}$, we have
\begin{align}
p(y=c|\bm{x}_q) = \frac{\exp{(-d(f_\theta(\bm{x}_q)W, \bm{p}_cW))}}{\sum_{\bm{p}_c\in\mathcal{P}}\exp{(-d(f_\theta(\bm{x}_q)W, \bm{p}_cW)}}
\end{align}
The distance function $d$ commonly adopts the cosine distance or squared Euclidean distance. Learning proceeds by minimizing the classification loss $\mathcal{L}_{cls}$, which is formatted as:
\begin{equation}
\begin{split}
    \mathcal{L}_{cls} &= \frac{1}{|\mathcal{Q}|}\sum_{\bm{x}_q\in\mathcal{Q}}[d(f_\theta(\bm{x}_q)W, \bm{p}_cW)\\&+\log\sum_{\bm{p}_c\in\mathcal{P}}\exp{(-d(f_\theta(\bm{x}_q)W, \bm{p}_cW))}]
\end{split}
\label{eq:cls}
\end{equation}

\subsection{Discriminative Reference Regularization}\label{sec:disc}
To further improve TART, we propose a Discriminative Reference Regularization (DRR) for more discriminative metric spaces. 
Since the transformation matrix is only decided by the reference layer and the prototype set of the given task, these task-independent reference vectors are the key elements to constructing discriminative metric spaces.
For training the reference vectors, we propose to maximize the distance between all transformed prototypes in the task-adaptive metric spaces during training. 
Different from contrastive learning, our DRR requires no additional data and focuses more on learning task-independent reference vectors instead of the feature encoder for the downstream task.
Formally, for a particular episode, given the prototype set $\mathcal{P}=\{\bm{p}_1,\dots,\bm{p}_{N}\}$ and the transformation matrix $W$, the discriminative loss $\mathcal{L}_{drr}$ is defined as:
\begin{align}
\mathcal{L}_{drr} = \sum_{i\neq j, \bm{p}\in \mathcal{P}}-d(\bm{p}_iW,\bm{p}_jW)
\label{eq:drr}
\end{align}
The total loss for training our TART model is thus $\mathcal{L} = \mathcal{L}_{cls} + \lambda\mathcal{L}_{drr}$, where $\lambda$ serves as regularization strength. 
Empirically, we set $\lambda=0.5$ in our experiments.



\begin{algorithm}[htb]
\small
\caption{TART Training Procedures}
\label{algo:train}
\begin{algorithmic}[1]
	\Require{A feature encoder $f_\theta$, a training set $\mathcal{D}_{train}=\{(\mathcal{S}_1,\mathcal{Q}_1),\dots,(\mathcal{S}_T,\mathcal{Q}_T)\}$, reference layers $\{\bm{r}_1,\dots,\bm{r}_N\}$.}
	\State Randomly initialize the model parameters and reference layers.
        \For{each episode $(\mathcal{S}_i,\mathcal{Q}_i)\in \mathcal{D}_{train}$}
            \State $\mathcal{L}_{cls} \gets 0$, $\mathcal{L}_{drr} \gets 0$
            \For{$k$ in $\{1,\dots,N\}$}
            \State $\bm{p}_c \gets \frac{1}{|\mathcal{S}_c|}\sum_{(\bm{x}_i,y_i)\in \mathcal{S}_c}f_{\theta}(\bm{x}_i)$
            \EndFor
            \State $R\gets[\frac{\bm{r}_1}{\parallel \bm{r}_1\parallel},\dots,\frac{\bm{r}_{N}}{\parallel \bm{r}_{N}\parallel}]$
            \State $P_i\gets[\frac{\bm{p}_1}{\parallel \bm{p}_1\parallel},\dots,\frac{\bm{p}_{N}}{\parallel \bm{p}_{N}\parallel}]$
            \State $W_i=\{P_i^{T}P_i\}^{-1}P_i^{T}R$
        \For{$k$ in $\{1,\dots,N\}$}
        \For{$(\bm{x},y)$ in $\mathcal{Q}_i$}
	 \State Compute $\mathcal{L}_{cls}$ using Eq.\ref{eq:cls}
        \EndFor	
        \EndFor
        \State Compute $\mathcal{L}_{drr}$ using Eq.\ref{eq:drr}
        \State Update model parameters minimizing $\mathcal{L}$ via optimizer
        \EndFor
\end{algorithmic}
\end{algorithm}

For better understanding, the whole training procedure for TART is summarized in Algorithm \ref{algo:train}.
The model parameters and reference layers are randomly initialized.
Given each training episode, we randomly chose $T$ episodes of the support set and query set from the training dataset, each episode consists of $K$ labeled samples over $N$ classes. Then, with the support set $\mathcal{S}_c$ for class $c$, the prototype $\bm{p}_c$ is obtained for each class (in line 5). 
Based on the prototype set and the reference layers, the transformation matrix $W$ is computed as a task-adaptive projection matrix (in lines 7-9). 
For each query input, the distances between the transformed query embeddings and each transformed prototype are measured in the task-adaptive metric space, and the classification loss $\mathcal{L}_{cls}$ is computed using these distances (in lines 10-12).
The discriminative loss is obtained over the prototype set for each episode (in line 15). 
The learnable parameters of the feature encoder and the reference layers are updated based on the total loss $\mathcal{L}$ (in line 16). 
This process gets repeated for every remaining episode with new classes of texts and queries.

\section{Experiments}

\begin{table*}[htb]
\small
\begin{center}
\begin{tabular}{ccccc}
\toprule
Dataset  & \# samples & Avg. \# tokens/sample & Vocab size & \# train/val/test classes \\ \hline
Huffpost & 36,900     & 11                      & 8,218      & 20/5/16                   \\
Amazon   & 24,000     & 140                     & 17,062     & 10/5/9                    \\   
Reuters  & 620        & 168                     & 2,234      & 15/5/11                   \\
20 Newsgroups   & 18,820     & 340                     & 32,137     & 8/5/7             \\
\bottomrule
\end{tabular}
\end{center}
\vspace{-0.3cm}
\caption{Statistics of the four benchmark datasets. \textit{Avg. \# tokens/sample} denotes the average tokens per sample.}
\label{tab:statitics}
\end{table*}

\subsection{Datasets}
We use four benchmark datasets for the evaluation of few-shot text classification task, whose statistics are summarized in Table~\ref{tab:statitics}.

\textbf{HuffPost headlines}  consists of news headlines published on HuffPost between 2012 and 2018~\cite{misra2018news}. These headlines are split into 41 classes. In addition, their sentences  are shorter and less grammatically correct than formal phrases.

\textbf{Amazon product data} contains product reviews from 24 product categories, including 142.8 million reviews spanning 1996-2014~\cite{he2016ups}. Our task is to identify the product categories of the reviews. Due to the huge size of the original dataset, we sample a subset of 1,000 reviews from each category.

\textbf{Reuters-21578} is collected from Reuters articles in 1987~\cite{lewis1997reuters}. We use the standard ApteMode version of the dataset. Following~\citet{bao2019few}, we evaluate 31 classes and eliminate articles with multiple labels. Each class comprises a minimum of twenty articles.

\textbf{20 Newsgroups} is a collection of approximately 20,000 newsgroup documents~\cite{lang1995newsweeder}, partitioned equally among 20 different newsgroups.

\subsection{Baselines.}
We compare our TART with multiple competitive baselines, which are briefly summarized as follows:
\begin{enumerate*}[label=(\roman*)]
    \item \textbf{MAML}~\cite{finn2017model} is trained by maximizing the sensitivity of the loss functions of new tasks so that it can rapidly adapt to new tasks once the parameters have been modified via a few gradient steps.
    \item \textbf{Prototypical Networks}~\cite{snell2017prototypical}, abbreviated as PROTO, is a metric-based method for few-shot classification by using sample averages as class prototypes.
    \item \textbf{Latent Embedding Optimization}~\cite{rusu2018meta},
    abbreviated as LEO, learns a low-dimensional latent embedding of model parameters and performs gradient-based meta-learning in this space.
    \item \textbf{Induction Networks}~\cite{geng2019induction} learns a class-wise representation by leveraging the dynamic routing algorithm in meta-learning.
    \item \textbf{HATT}~\cite{gao2019hybrid} extends PROTO by adding a hybrid attention mechanism to the prototypical network.
    \item \textbf{DS-FSL}~\cite{bao2019few} maps the distribution signatures into attention scores to extract more transferable features.
    \item \textbf{MLADA}~\cite{han2021meta} adopts adversarial networks to improve the domain adaptation ability of meta-learning.
    \item \textbf{Frog-GNN}~\cite{xu2021frog} extracts better query representations with multi-perspective aggregation of graph node neighbors.
    \item \textbf{P-Tuning}~\cite{liu2021gpt} is a prompt-based method that employs soft-prompting techniques to optimize prompts in continuous space.
    \item \textbf{LEA}~\cite{hong2022lea} determines the top-$k$ most relevant attention aspects to utilize pre-trained models in few-shot learning.
\end{enumerate*}

\subsection{Implementation Details}
In accordance with prior work~\cite{bao2019few}, we use pre-trained fastText~\cite{joulin2016fasttext} for word embedding.
As a feature extractor, we employ a BiLSTM with 128 hidden units and set the number of hidden units for the reference layers to 256.
We take cosine similarity as the distance function.
The model is implemented in PyTorch~\cite{paszke2017automatic} using the Adam~\cite{kingma2014adam} optimizer with a $10^{-4}$ learning rate.
For the sake of a fair comparison, we follow the identical evaluation protocol and train/val/test split as ~\citet{bao2019few}. The model parameters and reference layers are randomly initialized. During meta-training, we perform 100 training episodes per epoch. Meanwhile, we apply early stopping if the accuracy on the validation set does not increase after 20 epochs. 
We evaluate the model performance based on a total of one thousand testing episodes and present the average accuracy across five different random seeds. 
All experiments are conducted with NVIDIA V100 GPUs.

\subsection{Comparisons}
\begin{table*}[htb]
\resizebox{\textwidth}{!}{
\begin{tabular}{ccccccccccccccc}
\toprule
\multirow{2}{*}{Method} & \multicolumn{2}{c}{HuffPost}  &  & \multicolumn{2}{c}{Amazon}    &  & \multicolumn{2}{c}{Reuters}   &  & \multicolumn{2}{c}{20 News}   &  & \multicolumn{2}{c}{Average}   \\ \cline{2-15}
                        & 1 shot        & 5 shot        &  & 1 shot        & 5 shot        &  & 1 shot        & 5 shot        &  & 1 shot        & 5 shot        &  & 1 shot        & 5 shot        \\ \hline
MAML (2017)                    & 35.9          & 49.3          &  & 39.6          & 47.1          &  & 54.6          & 62.9          &  & 33.8          & 43.7          &  & 40.9          & 50.8          \\
PROTO (2017)                   & 35.7          & 41.3          &  & 37.6          & 52.1          &  & 59.6          & 66.9          &  & 37.8          & 45.3          &  & 42.7          & 51.4          \\
LEO* (2018)                     & 28.8          & 42.3          &  &  39.5         & 52.5          &  & 35.4          & 54.1          &  & 36.4          & 52.2          &  & 35.0          &  50.3         \\
Induct (2019)                  & 38.7          & 49.1          &  & 34.9          & 41.3          &  & 59.4          & 67.9          &  & 28.7          & 33.3          &  & 40.4          & 47.9          \\
HATT (2019)                    & 41.1          & 56.3          &  & 49.1          & 66.0          &  & 43.2          & 56.2          &  & 44.2          & 55.0          &  & 44.4          & 58.4          \\
DS-FSL (2020)                  & 43.0          & 63.5          &  & 62.6          & 81.1          &  & 81.8          & 96.0            &  & 52.1          & 68.3          &  & 59.9          & 77.2          \\
MLADA (2021)                   & 45.0          & 64.9          &  & 68.4          & \textbf{86.0} &  & 82.3          & 96.7          &  & 59.6          & 77.8          &  & 63.9          & 81.4          \\ 
LEA (2022)                  & 46.2          & 65.8          &  & 66.5          & 83.5          &  & 69.0       & 89.0            &  & 54.1          & 60.2          &  & 58.9          & 74.6        \\ \hline
TART w/o DRR                   & \textbf{48.4} & 66.0          &  & 68.9          & 83.5          &  & 90.4          & 96.2          &  & 66.4          & 82.2          &  & 68.5          & 81.9          \\
TART              & 46.9          & \textbf{66.8} &  & \textbf{70.1} & 82.4 &  & \textbf{92.2} & \textbf{96.7} &  & \textbf{67.0} & \textbf{83.2} &  & \textbf{69.0} & \textbf{82.3} \\ \bottomrule
\end{tabular}}
\caption{Results of 5-way 1-shot and 5-way 5-shot classification on four datasets. The bottom two rows present our ablation study. *Reported by ~\citet{hong2022lea}.}
\label{tab:results}
\end{table*}

\begin{table*}[htb]
\resizebox{\textwidth}{!}{
\centering
\begin{tabular}{ccccccccccccc}
\toprule
\multirow{2}{*}{Method} & \multirow{2}{*}{PLM} & \multirow{2}{*}{EK} & \multicolumn{2}{c}{HuffPost}   & \multicolumn{2}{c}{Amazon}      & \multicolumn{2}{c}{Reuters}     & \multicolumn{2}{c}{20 News}     & \multicolumn{2}{c}{Average}     \\ \cline{4-13}
                        &                         &                     & 1-shot         & 5-shot        & 1-shot         & 5-shot         & 1-shot         & 5-shot         & 1-shot         & 5-shot         & 1-shot         & 5-shot         \\ \hline
LEA                     & $\times$                        &  $\times$                   & 48.4          & \textbf{71.6}   & 63.6           & 82.7          & 71.6          & 83.1          & 53.5          & 65.9          & 59.3          & 75.8          \\ 
Frog-GNN                     & $\times$                        &  $\times$                   & 54.1          & 69.6       & 71.5           & 83.6          & -          & -          & -          & -          & -          & -          \\ 
P-Tuning                & $\checkmark$                        &  $\times$                   & \textbf{54.5}          & 65.8         & 62.2          & 79.1          & \textbf{90.0}          & \textbf{96.7} & 56.2          & 77.7          & 65.7          & 79.8          \\ 
ContrastNet             &  $\times$                       &   $\checkmark$                  & 53.1      & 65.3         & \textbf{76.1}      & \textbf{85.2}         & 86.4          & 95.3          & 71.7          & 81.6          & \textbf{71.8}          & 81.9          \\ \hline
TART                    &  $\times$                       &  $\times$                   & 46.5          & 68.9         & 73.7          & 84.3          & 86.9   & 95.6          & \textbf{73.2} & \textbf{84.9} & 70.1 & \textbf{83.4} \\ \bottomrule
\end{tabular}}
\caption{5-way 1-shot and 5-way 5-shot classification on four datasets using BERT. \textit{PLM} denotes prompting language model and \textit{EK} denotes extra knowledge. Note that \textit{ContrastNet} utilizes a pre-trained short-texts paraphrasing model to generate data augmentation of texts.}
\label{tab:bert}
\end{table*}

\begin{figure*}[htb]
\centering  
\subfigure[AVG]{
\label{fig:tsnesub1}
\includegraphics[width=0.32\textwidth]{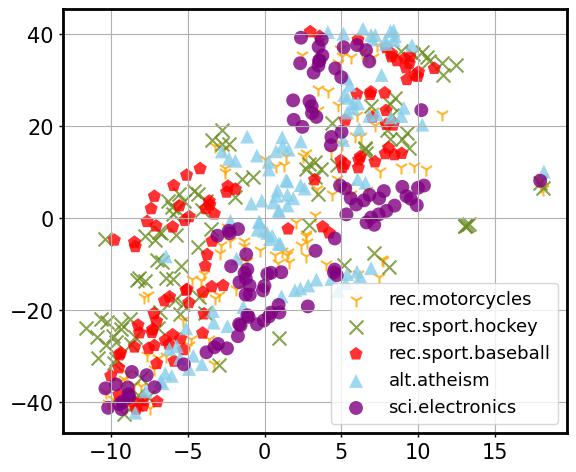}}
\subfigure[MLADA]{
\label{fig:tsnesub2}
\includegraphics[width=0.32\textwidth]{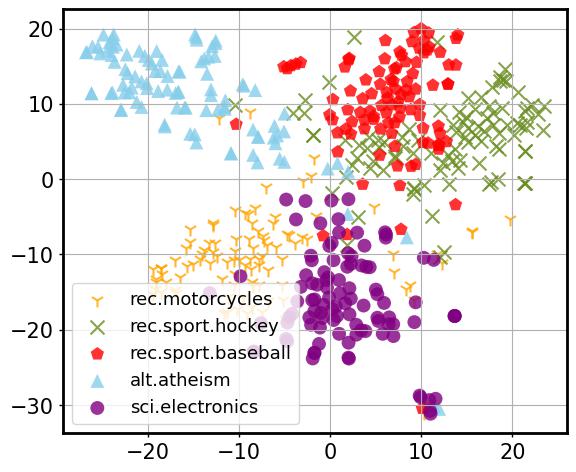}}
\subfigure[TART]{
\label{fig:tsnesub3}
\includegraphics[width=0.32\textwidth]{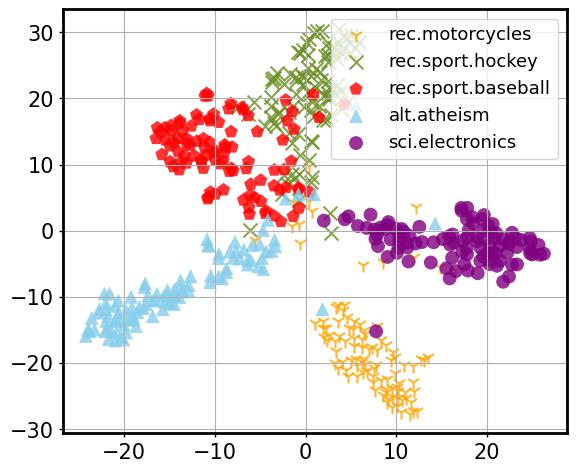}}
\caption{t-SNE visualization of the input representation of the classifier for a testing episode (N = 5, K = 5, Q = 100) sampled from 20 Newsgroups. Note that the 5 classes are not seen in training set. The input representation of the classifier is given by (a) the average of word embeddings (b) MLADA and (c) TART (ours).}
\label{fig:tsne}
\end{figure*}

The experimental results are shown in Table \ref{tab:results} in terms of various datasets, methods, and few-shot settings.
As demonstrated in Table \ref{tab:results}, our model outperforms recent methods across all datasets, with the exception of Amazon's 1-shot setting. 
In particular, our model achieves an average accuracy of 69.0\% for 1-shot classification and 82.3\% for 5-shot classification. 
Our model surpasses the state-of-the-art approach MLADA~\cite{han2021meta} by an average of 5.1\% in 1-shot and 0.9\% in 5-shot, demonstrating the effectiveness of task-adaptive metric space.
Specifically, our method delivers a substantial improvement, 9.9\% in 1-shot on Reuters, and 7.4\% and 5.4\% in 1-shot and 5-shot on 20Newsgroup, respectively.
The average length of texts in these datasets is longer than in the other datasets, verifying its superiority in the longer texts.
Moreover, we show that our model achieves a more significant boost in the 1-shot than in the 5-shot, indicating that our model contributes more to a generation of distinguishable class representation, particularly when the labeled class sample is limited.
\subsection{Ablation Study}
We conduct extensive studies to examine the effects of DRR, contextualized representations and reference vectors.

First, we study how the DRR affects the performance of our model. The results are presented at the bottom of Table~\ref{tab:results}. With the use of DRR, the model can construct a more discriminative subspace for classification, especially in 1-shot settings. This empirical study validates the effectiveness of DRR in enhancing performance.

We also experiment with contextualized representations, given by the pure pre-trained \texttt{bert-base-uncased} model, dubbed BERT$_{BASE}$~\cite{devlin2018bert}. The results are shown in Table~\ref{tab:bert}. We observe that BERT improves classification performance for the text-level dataset. Even while ContrasNet requires a pre-trained short-texts paraphrasing model to generate data augmentation, our model can outperform it without requiring any additional knowledge on the 5-shot setting.
\begin{table*}[ht]
\begin{center}
\resizebox{\textwidth}{!}{
\begin{tabular}{ccccccccccc}
\toprule
\multirow{2}{*}{Method} & \multicolumn{2}{c}{HuffPost} & \multicolumn{2}{c}{Amazon} & \multicolumn{2}{c}{Reuters} & \multicolumn{2}{c}{20 News} & \multicolumn{2}{c}{Average} \\ \cline{2-11} 
                        & 1-shot        & 5-shot       & 1-shot       & 5-shot      & 1-shot       & 5-shot       & 1-shot       & 5-shot       & 1-shot       & 5-shot       \\ \hline
1 layer Bi-LSTM         & 45.0          & 64.9         & 68.4         & \textbf{86.0}        & 82.3         & 96.7         & 59.6         & 77.8         & 63.9         & 81.4         \\
2 layer Bi-LSTM         & 45.2          & 65.2         & 67.1         & 83.7        & 85.5         & 96.4         & 64.0         & 78.6         & 65.5         & 81.0         \\
3 layer Bi-LSTM         & 45.4          & 63.6         & 66.0         & 83.2        & 84.3         & \textbf{97.9}         & 64.4         & 78.5         & 65.0         & 80.8         \\ \hline
TART                    & \textbf{46.9}          & \textbf{66.8}         & \textbf{70.1}         & 82.4        & \textbf{92.2}         & 96.7         & \textbf{67.0}         & \textbf{83.2}         & \textbf{69.0}         & \textbf{82.3}         \\ \bottomrule
\end{tabular}}
\caption{Comparison of the feature encoder with different numbers of layers.}
\label{tab:layers}
\end{center}
\end{table*}

The introduction of the reference vectors is to enhance the divergence between class prototypes in the metric space. Even though adding more layers to the feature encoder could theoretically make it better, the small number of labeled samples is probably causing it to overfit. Moreover, we investigate the performance of the feature encoder with multiple layers. We adopt MLADA as the basic model, which leads to state-of-the-art performance. The results are shown in Table~\ref{tab:layers}. We found that the feature encoder with two layers of Bi-LSTM achieves better performance than one layer in a 1-shot setting. But the accuracy decreases when the number of layers increases further.
In contrast, our model uses a linear transformation matrix that is figured out by using the reference layer and the prototypes of the support set. This can effectively enhance generalization while avoiding overfitting since it introduces fewer learnable parameters.

\begin{figure*}[htb]
    \centering
    \scalebox{0.95}{\includegraphics[width=\linewidth]{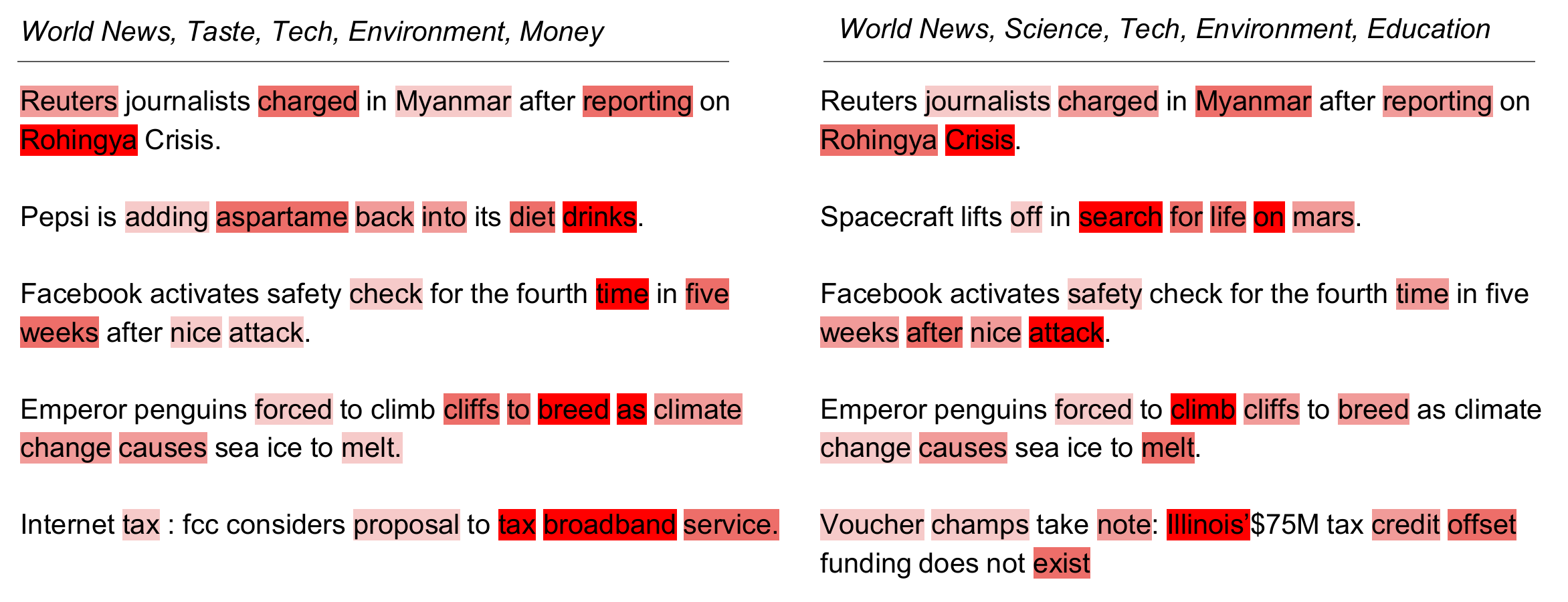}}
    \caption{The visualization of task-specific attention weights generated by our model.
    We visualize our model’s support sets of two different tasks (5-way 1-shot) in Huffpost dataset. 
     Word ``Crisis" is downweighed for \textit{World News} class when compared to \textit{Taste}, \textit{Tech}, \textit{Environment} and \textit{Money} classes (left), but it becomes important when replacing \textit{Taste} and \textit{Money} with \textit{Science} and \textit{Education} (right).
    }
    \label{fig:att}
\end{figure*}
\subsection{Hyperparameter Analysis}
\begin{table}[htb]
\small
\begin{center}
\begin{tabular}{c|cc|cc}
\toprule
\multirow{2}{*}{Settings} & \multicolumn{2}{c|}{Reuters}                       & \multicolumn{2}{c}{20 News}                      \\ \cline{2-5} 
                          & \multicolumn{1}{c|}{1-shot}        & 5-shot        & \multicolumn{1}{c|}{1-shot}      & 5-shot        \\ \hline
$\lambda$ = 0.3                & \multicolumn{1}{c|}{91.5}          & 96.3          & \multicolumn{1}{c|}{66.6}        & 82.9          \\
$\lambda$ = 0.5                & \multicolumn{1}{c|}{\textbf{92.2}} & \textbf{96.7} & \multicolumn{1}{c|}{\textbf{67.0}} & \textbf{83.2} \\
$\lambda$ = 0.7                & \multicolumn{1}{c|}{89.5}          & 95.4          & \multicolumn{1}{c|}{66.1}        & 82.0          \\
$\lambda$ = 0.9                & \multicolumn{1}{c|}{89.1}          & 94.9          & \multicolumn{1}{c|}{65.7}        & 81.7          \\ \bottomrule
\end{tabular}
\caption{Evaluation accuracy on the validation set of Reuters and 20 Newsgroup datasets. Different settings adjust the proportion of $\mathcal{L}_{drr}$.}
\label{tab:hyper}
\vspace{-0.3cm}
\end{center}
\end{table}

We analyze the effect of different settings of hyperparameter $\lambda$. Table \ref{tab:hyper} demonstrates the accuracy in different settings on the validation set of the Reuters and 20 Newsgroup datasets. 
We discover that $\lambda=0.5$ yields the optimum performance, and further reduction/increase in the ratio lead to performance degradation. It is likely because $\mathcal{L}_{drr}$ can improve the divergence of the class prototypes. But a too-large ratio of $\mathcal{L}_{drr}$ would make the model focus more on the task-independent reference vectors while ignoring the learning for a unique feature space, which may lead to an over-fitting problem.


\subsection{Visualization} 

We utilize visualization experiments to demonstrate that our model can build high-quality sentence embeddings and identify significant lexical features for unseen classes.

To illustrate that our model can generate high-quality sentence embeddings for unseen classes, we view the high-dimensional features as two-dimensional images using the t-SNE algorithm~\cite{van2008visualizing}. Figure~\ref{fig:tsne} depicts the 256-dimension feature representations for a 5-way 5-shot testing episode sampled from the 20 NewsGroup dataset. 
From the results, it is evident that the distances between the inter-classes are much larger than those of the average word embeddings and MLADA depicted in Figure~\ref{fig:tsnesub1} and Figure~\ref{fig:tsnesub2}, respectively.
This enlarged inter-class spacing shows that our method can construct a more distinct feature space for each episode.

In addition, the weight vectors on the same support samples are depicted in two testing episodes. The example is drawn from the Huffpost dataset. Figure~\ref{fig:att} demonstrates that our apporach is capable of generating task-specific attention. Even with the same text, the attention of each word varied based on the different combinations of categories in the task. Specifically, as compared to \textit{Science} and \textit{Education} class, Word ``Crisis", ``attack" and ``climb" become more important for \textit{World News}, \textit{Tech} and \textit{Education} class, respectively.

\section{Conclusion}
In this work, we propose a novel TART for few-shot text classification, which can enhance the generalization by transforming the class prototypes to per-class fixed reference points in task-adaptive metric spaces.
Specifically, a task-adaptive transfer module is designed to project embedding features into a task-specific metric space by using a linear transformation matrix.
In addition, we propose a discriminative reference regularization to maximize divergence between transformed prototypes in task-adaptive metric spaces.
The proposed model is evaluated on four standard text classification datasets.
Without any extra knowledge or data information, our TART outperforms previous work by a large margin.

\section{Limitations}
Our approach is based on meta-learning and is designed for constrained situations where computing resources are limited, such as on-device settings. Therefore, using large and complex feature encoders like LLM may pose scalability challenges.
In addition, if the task involves a significant number of new classes, the model may not scale effectively.
Lastly, our method is primarily suitable for text classification, such as news category or product review classification. It is not appropriate for text generation tasks.


\bibliography{custom}
\bibliographystyle{acl_natbib}

\clearpage

\end{document}